\documentclass[conference]{IEEEtran}
\IEEEoverridecommandlockouts
\usepackage{cite}
\usepackage{amsmath,amssymb,amsfonts}
\usepackage{algorithmic}
\usepackage{graphicx}
\usepackage{textcomp}
\usepackage{xcolor}
\def\BibTeX{{\rm B\kern-.05em{\sc i\kern-.025em b}\kern-.08em
    T\kern-.1667em\lower.7ex\hbox{E}\kern-.125emX}}

\begin{document}

\title{Classifier-Guided Captioning Across Modalities\\

}

\author{
\IEEEauthorblockN{
\begin{minipage}{0.2\textwidth}
\centering
Ariel Shaulov \\
\textit{Tel Aviv University}\\
arielshaulov\\@mail.tau.ac.il
\end{minipage}
\hfill
\begin{minipage}{0.2\textwidth}
\centering
Tal Shaharabany \\
\textit{Tel Aviv University}\\
shaharabany\\@mail.tau.ac.il
\end{minipage}
\hfill
\begin{minipage}{0.2\textwidth}
\centering
Eitan Shaar \\
\textit{Bar-Ilan University}\\
shaarei@biu.ac.il
\end{minipage}
%}
%\\
%\\
%\IEEEauthorblockN{
\begin{minipage}{0.2\textwidth}
\centering
Gal Chechik \\
\textit{Bar-Ilan University}\\
\textit{NVIDIA, Israel} \\
gal.chechik@biu.ac.il
\end{minipage}
\hfill
\begin{minipage}{0.2\textwidth}
\centering
Lior Wolf \\
\textit{Tel Aviv University}\\
wolf@cs.tau.ac.il
\end{minipage}
}
}
\maketitle

\begin{abstract}
Most current captioning systems use language models trained on data from specific settings, such as image-based captioning via Amazon Mechanical Turk, limiting their ability to generalize to other modality distributions and contexts. This limitation hinders performance in tasks like audio or video captioning, where different semantic cues are needed. Addressing this challenge is crucial for creating more adaptable and versatile captioning frameworks applicable across diverse real-world contexts. In this work, we introduce a method to adapt captioning networks to the semantics of alternative settings, such as capturing audibility in audio captioning, where it is crucial to describe sounds and their sources. Our framework consists of two main components: (i) a frozen captioning system incorporating a language model (LM), and (ii) a text classifier that guides the captioning system. The classifier is trained on a dataset automatically generated by GPT-4, using tailored prompts specifically designed to enhance key aspects of the generated captions. Importantly, the framework operates solely during inference, eliminating the need for further training of the underlying captioning model. We evaluated the framework on various models and modalities, with a focus on audio captioning, and report promising results. Notably, when combined with an existing zero-shot audio captioning system, our framework improves its quality and sets state-of-the-art performance in zero-shot audio captioning.
\end{abstract}


\section{Introduction}
Captioning is the task of generating descriptive text for modalities like images or audio. It has numerous applications from information retrieval to accessibility, but is fundamentally challenging because it may require deep semantic understanding of the content.  
Current baseline models rely on language models (LMs) as the core component for caption generation. However, the use of LMs presents challenges for certain modalities due to shifts in data distribution, as LMs are trained primarily on textual data and thus struggle to adapt effectively to modality-specific captioning tasks.

% In audio captioning, it is crucial to generate captions that produce audible effects. Distinguishing between audible actions, such as speaking or clapping, and non-audible ones, like walking, is essential for generating accurate and contextually relevant caption.

{In audio captioning, it is crucial to generate captions that capture nuanced audio-specific semantics, such as distinguishing between audible actions, like speaking or clapping, and non-audible ones, like walking. Additionally, models often struggle to adapt to diverse real-world contexts, where varying environmental and cultural factors influence audio interpretation and captioning requirements, making accurate and contextually relevant captions more challenging to generate.}

To differentiate between audible and non-audible actions, we trained a text classifier on a dataset automatically generated by directing GPT-4. This classifier is employed during inference to guide the LM, allowing it to select more audible words in the caption generation process.

We validate the effectiveness of our framework through a series of experiments, particularly in the domain of audio captioning. Our method is evaluated on the AudioCaps \cite{kim2019audiocaps} and Clotho \cite{drossos2020clotho} datasets, where we demonstrate that incorporating audibility guidance significantly enhances performance across all metrics compared to baseline models. Additionally, we conducted an ablation study to show the applicability of our approach to other modalities, such as image captioning. 

Our contributions can be summarized as follows: (1) We propose a novel approach to adapt captioning networks to the semantics of alternative settings by incorporating classifier guidance, (2)  We introduce a carefully curated dataset, generated by GPT-4, containing both audible and non-audible caption samples, and (3) We demonstrate that our framework achieves state-of-the-art results in zero-shot audio captioning and significantly enhances performance in image captioning across both zero-shot and fully supervised settings.

\begin{figure*}[t]
\centerline{\includegraphics[width=\textwidth]{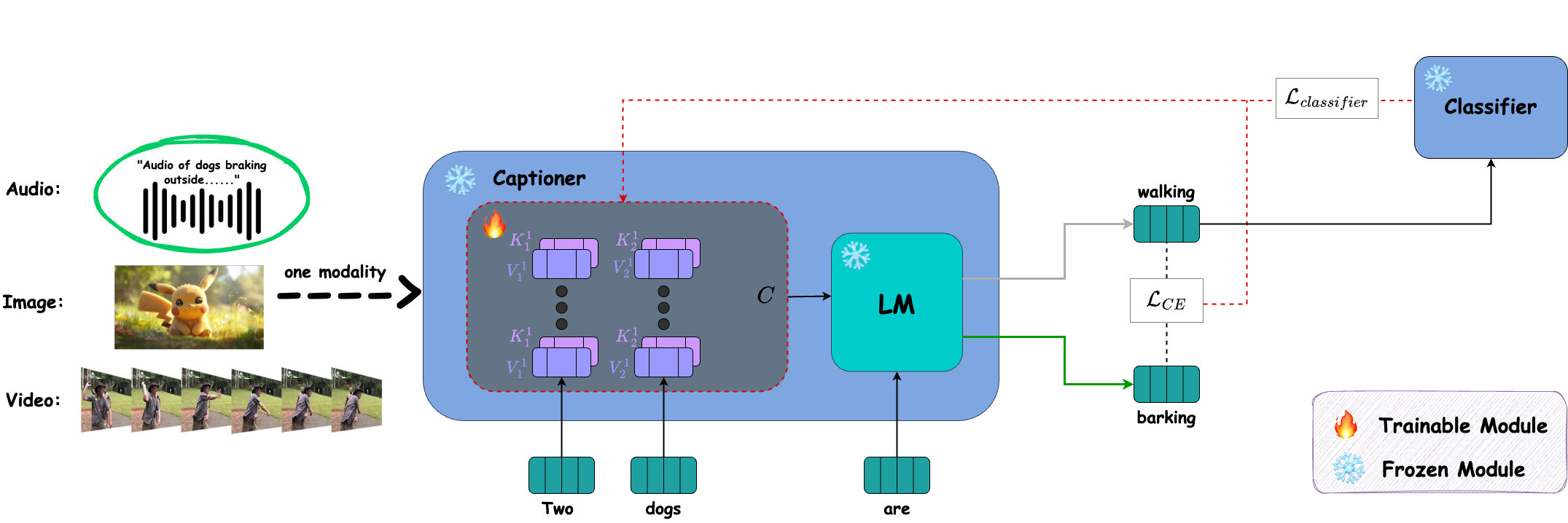}}
\caption{
An overview of our modality-agnostic approach is depicted, with a focus on audio captioning during inference. In this illustration, the audio modality is highlighted (enclosed in a green circle), where the captioner functions as an audio captioner, and the classifier is an audibility classifier (which was pre-trained on our audibility dataset, see Sec.~\ref{Audibility classifier}). The model is guided to generate the phrase 'barking' instead of 'walking' by adjusting the context (C) using the gradients from the classifier loss \(L_{classifier}\), as indicated by the red arrow. To maintain the inherent characteristics of the language model, the minimum divergence from the original distribution is optimized using the cross-entropy loss \(L_{ce}\).}
\label{architecture}
\end{figure*}

\section{Related Work}
Captioning models, which generate descriptive text for input modalities, have been widely studied in artificial intelligence. Traditional methods focused on image captioning, using convolutional neural networks (CNNs) for feature extraction and recurrent neural networks (RNNs) or transformers for sequence generation \cite{vinyals2015show, karpathy2015deep, donahue2015long}, with improvements from attention mechanisms \cite{xu2015show, you2016image, anderson2018bottom}.

Audio captioning, which generates descriptions for audio inputs, initially adapted techniques from image captioning, using spectrogram-based CNNs combined with RNNs or transformers \cite{drossos2017automated, koizumi2020audio}. Recent advancements introduced attention mechanisms tailored to audio data, improving contextual relevance \cite{kim2019audiocaps, wu2019audio}.

% Zero-shot methods like NoAudioCaptioning \cite{deshmukh2024training}, WSAC \cite{kouzelis2023weakly}, and Zhang et al. \cite{zhang2024zero} propose using text-only data for training, bypassing the need for audio inputs and leveraging pre-trained language models to simulate training.

% Despite these advancements, challenges remain in capturing audibility, which is the ability to describe (only the) elements inferred from the audio. 

{Zero-shot methods like NoAudioCaptioning \cite{deshmukh2024training}, WSAC \cite{kouzelis2023weakly}, and Zhang et al. \cite{zhang2024zero} propose using text-only data for training, bypassing the need for audio inputs and leveraging pre-trained language models to simulate training. While these methods achieve impressive results, they lack modality-specific guidance, which is crucial for accurately capturing nuanced audio features. Specifically, these approaches struggle with capturing audibility—the ability to describe only the elements directly inferred from the audio—due to their reliance on text-based training that overlooks the unique characteristics and dynamics of audio data.  In contrast, our inference-time classifier provides audibility-focused optimization without retraining the language model,
offering both adaptability and efficiency in zero-shot and supervised settings}

\section{Method}
The task of captioning can be mathematically formulated as a sequence generation problem, where we aim to infer the conditional probability of the $i$-th word, denoted as $x_i$, in a sentence. Specifically, the objective is to optimize the probability distribution $P(x_i|[x_t]_{t < i}, \mathcal{A})$, where $x_t$ represents the preceding words in the sentence, and $\mathcal{A}$ denotes the input, in our case, an audio clip.

In this work, we focus on zero-shot audio-captioning by introducing a novel inference-time optimization method designed to generate high-fidelity, audibly-relevant captions. This approach leverages audibility classifier-based guidance to direct a language model (LM) towards producing more audibly interpretable outputs.

Notably, our method is modality-agnostic, allowing seamless integration across various modalities such as audio and images, thus offering broad applicability and flexibility.

The proposed framework consists of two main components: (i) a pre-trained audio captioner, which incorporates a language model (LM), and (ii) a binary audibility classifier that provides guidance during the caption generation process.

The sentence generation process is mathematically expressed as:
\[
x_{i+1} = \operatorname{LM}\left( x_i, [(K_j^l, V_j^l)]_{j < i, 1 \leq l \leq L} \right),
\]
where $x_i$ denotes the $i$-th word in the generated sentence, and $K_j^l$ and $V_j^l$ represent the context transformer's key and value for the $j$-th token across $L$ layers.

To enhance the captioner's ability to generate sentences that reflect the audibility of the audio inputs, we introduce a calibrated audibility loss, $L_{classifier}$, which encourages the model to produce more audibly accurate sentences. This is achieved by modifying, during inference, the context cache values $C_i = \left[ (K^l_j, V^l_j) \right]_{j < i, 1 \leq l \leq L}$, while keeping the LM itself unchanged.

Additionally, we incorporate a cross-entropy loss, $L_{ce}$, to ensure that the distribution of the next token remains consistent with that of the original language model.

The optimization process occurs during inference through $n$ optimization steps, where the next token is chosen iteratively to refine the model's output. In this dual-loss framework, the objective is to balance the generation of audibly descriptive captions while maintaining the fluency and coherence inherent to the language model. We illustrate our framework in Fig.~\ref{architecture}.

\begin{table}[t]
\caption{Examples from the GPT-4 Generated Dataset Comprising Both Audible and Non-Audible Sentences. We Prompted ChatGPT with the Task of Creating Examples for a Classifier that Distinguishes Between These Two Categories. Note: The Examples in the Table Are Not Meant to Match; the Two Lists Are Separate.}
\begin{center}
\renewcommand{\arraystretch}{1.2} % Adjust row height
\setlength{\tabcolsep}{4pt} % Adjust column width
\scriptsize % Reduce font size
\begin{tabular}{|p{0.45\linewidth}|p{0.45\linewidth}|}
\hline
\textbf{Audible} & \textbf{Non-Audible} \\
\hline
The barking of a dog in excitement. & Magnets attract metals.\\
Ringing phone awaits an answer. & Ice covers the lake in winter.\\
The buzz of a drone flying overhead. & Icebergs float on water. \\
Jingling coins are counted or played with. & A statue in a park.\\
Whips cracked in the rodeo. & Resolved issue is fixed.\\
The meow of a cat. & Rusting car sits in the yard.\\
\hline
\end{tabular}
\label{tab1}
\end{center}
\end{table}

\subsection{Audibility Classifier}\label{Audibility classifier}
To address the challenge of insufficient audibility in sentences generated by audio captioning models, we generated two distinct sets of sentences using ChatGPT: one set representing audible captions and the other representing non-audible captions. This newly generated dataset serves as the foundation for training a classifier ($h_a$), specifically designed to distinguish between audible and non-audible captions.

In Table~\ref{tab1}, we provide examples of sentences from the dataset used for training the classifier to distinguish between audible and not audible descriptions. These sentences were generated using the following prompt to ChatGPT-4:
\begin{quote}
``I want to train a classifier that distinguishes between an audio description that is more audible and non-audible - two classes. generate examples for each category. You should consider factors such as the coherence of words, grammatical correctness, context, and the likelihood that the sentence represents a meaningful auditory scenario.''.
\end{quote}

The trained classifier $h_a$ offers guidance to the Language Model (LM) in a manner that aligns with our audibility objectives. This guidance is attained through optimizing the following term over the context cache $C_i$:

\begin{equation}
\label{eq:one}
L_{classifier} = -\log\left(h_a(LM(x_i;C_i)[1]\right)
\end{equation}
Where $[1]$ indexes the classifier's output for the pseudo-probability for the audibility label (the positive label). {The audibility classifier evaluates the sentence in its entirety, and accordingly, the guidance pertains to the sentence as a whole.}

\subsection{Loss Function}
Our methodology employs a pre-trained audio captioner that contains a LM (GPT-2 in our case) to deduce the subsequent word in a sentence.

To ensure that the captioner generates sentences that are influenced by the classifier, we incorporate an additional term, $L_{classifier}$, from Eq.~\ref{eq:one} into the main loss function.

Furthermore, an additional regularization term denoted as $L_{CE}$ (Eq.~\ref{eq:two}) is added (as is often done) 
% \eitan{where is it done? we need to cite the papers} 
to ensure that the distribution of the next token remains consistent with that of the original language model.

\begin{equation}
\label{eq:two}
L_{CE} = CE(LM(x_i;C_i),LM(x_i;C^o_i))
\end{equation}
where $i$ is the index of the currently generated token, CE is the cross entropy loss, and $C^o_i$ is a context cache of the relevant keys, queries, and values as they are computed based on the embedding and $K$ $Q$ and $V$ projections of the LM model (without the inference time optimization over $C_i$).

To conclude, the loss function of the optimization process can be represented as:
\begin{equation}
\label{eq:four}
\mathcal{L} = \lambda_0\cdot\mathcal{L}_{CE} + \lambda_1\cdot\mathcal{L}_{classifier}
\end{equation}

as default parameters, we set $\lambda_0$ to be 0.2, and $\lambda_1$ to be 0.6.

This optimization process is executed iteratively during auto-regression, with each token being addressed in sequence. At every generation step, we optimize the current context cache $C_i$ using gradient descent, generate the next token, and continue to the next iteration. {Importantly, this process does not involve any updates to the model weights, which remain fixed throughout.}

\begin{table*}[t]
\centering
\caption{The experimental results for out-of-domain scenarios on the AudioCaps and Clotho datasets. Aud = Audibility accuracy in percents. NAC = NoAudioCaptioning. (*) The results as originally presented in the papers.}
\renewcommand{\arraystretch}{1.2}
\setlength{\tabcolsep}{2pt}
\scriptsize
\begin{tabular}{|c|c|c|c|c|c|c|c|c|c|c|c|c|c|c|c|c|}
\hline
\textbf{Method} & \multicolumn{8}{|c|}{\textbf{AudioCaps $\Rightarrow$ Clotho}} & \multicolumn{8}{|c|}{\textbf{Clotho $\Rightarrow$ AudioCaps}} \\
\cline{2-17}
 & \textbf{BLEU\textsubscript{4}} & \textbf{ROUGE\textsubscript{L}} & \textbf{METEOR} & \textbf{SPICE} & \textbf{CIDEr} & \textbf{Aud} & \textbf{CLAP-S} & \textbf{BERT-S} & \textbf{BLEU\textsubscript{4}} & \textbf{ROUGE\textsubscript{L}} & \textbf{METEOR} & \textbf{SPICE} & \textbf{CIDEr} & \textbf{Aud} & \textbf{CLAP-S} & \textbf{BERT-S} \\
\hline
\multicolumn{17}{|c|}{\textbf{Fully Supervised Methods}} \\
\hline
Prefix AAC \cite{kim2023prefix}* & 6.5 & 27.6 & 11.2 & 7.4 & 19.2 & - & - & - & 8.4 & 33.0 & 14.4 & 8.3 & 21.1 & - & - & - \\
\hline
RECAP \cite{ghosh2024recap}* & 6.8 & 27.6 & 11.0 & 8.4 & 19.5 & - & - & - & 6.5 & 28.1 & 11.2 & 13.6 & 19.1 & - & - & - \\
\hline
\multicolumn{17}{|c|}{\textbf{Zero-Shot Methods}} \\
\hline
ZerAuCap \cite{salewski2023zeroshot}* & 2.9 & 25.4 & 9.4 & 5.3 & 14.0 & - & - & - & 6.8 & 31.1 & 12.3 & 8.6 & 28.1 & - & - & - \\
\hline
WSAC \cite{kouzelis2023weakly}* & - & 26.6 & 12.0 & 8.2 & 20.6 & - & - & - & - & 35.5 & 17.3 & 12.0 & 25.6 & - & - & - \\
\hline
Zhang et al \cite{zhang2024zero}* & - & 29.8 & 13.2 & 9.3 & \textbf{24.8} & - & - & - & - & 36.1 & 18.0 & 12.5 & \textbf{33.8} & - & - & - \\
\hline
NAC \cite{deshmukh2024training} & 7.0 & 28.6 & 12.7 & 8.7 & 21.3 & 59.8 & 0.75 & 0.80 & 12.2 & 35.5 & 17.7 & 11.6 & 27.8 & 65.6 & 0.79 & 0.85 \\
\hline
NAC \cite{deshmukh2024training} + Ours & \textbf{7.7} & \textbf{30.0} & \textbf{13.5} & \textbf{9.9} & 22.3 & \textbf{78.2} & \textbf{0.84} & \textbf{0.88} & \textbf{13.5} & \textbf{36.7} & \textbf{18.2} & \textbf{13.6} & 29.3 & \textbf{85.7} & \textbf{0.88} & \textbf{0.91} \\
\hline
\end{tabular}
\label{cross_domain_table}
\end{table*}

\begin{table}[t]
\caption{image captioning results on MS-COCO dataset}
\centering
\scriptsize
\setlength{\tabcolsep}{3pt}
\renewcommand{\arraystretch}{1.2}
\begin{tabular}{|c|c|c|c|c|c|}
\hline
\textbf{Method} & \multicolumn{4}{|c|}{\textbf{Supervised Metrics}} & \textbf{Unsupervised Metrics} \\
\cline{2-6}
 & \textbf{BLEU\textsubscript{4}} & \textbf{METEOR} & \textbf{CIDEr} & \textbf{SPICE} & \textbf{CLIP-S} \\
\hline
\multicolumn{6}{|c|}{\textbf{Fully Supervised Methods}} \\
\hline
ClipCap~\cite{mokady2021clipcap} & 32.1 & 27.1 & 108.3 & 20.1 & 0.77 \\
\hline
ClipCap~\cite{mokady2021clipcap} + ours & \textbf{33.2} & \textbf{27.7} & \textbf{110.2} & \textbf{20.8} & \textbf{0.81} \\
\hline
\multicolumn{6}{|c|}{\textbf{Zero-Shot Methods}} \\
\hline
ZeroCap~\cite{tewel2022zerocap} & 2.6 & 11.5 & 14.6 & 5.5 & 0.87 \\
\hline
ZeroCap~\cite{tewel2022zerocap} + ours & \textbf{3.2} & \textbf{12.5} & \textbf{15.2} & \textbf{5.9} & \textbf{0.89} \\
\hline
\end{tabular}
\label{ablation_table}
\end{table}

\section{Experiments}
The evaluation of audio captioning models heavily relies on high-quality datasets that provide paired audio and text data. Two widely recognized datasets in this domain are AudioCaps \cite{kim2019audiocaps} and Clotho \cite{drossos2020clotho}.

AudioCaps \cite{kim2019audiocaps}, derived from the AudioSet ontology, is a large-scale dataset specifically curated for audio captioning tasks. It comprises over 46k audio clips and 49k human-generated captions, each describing the acoustic events present in the corresponding audio.

In contrast, Clotho \cite{drossos2020clotho} offers a more diverse and challenging set of audio-caption pairs. The Clotho dataset contains nearly 5k audio files and 19k captions, with each audio file ranging from {15 to 30} seconds in length.

We train our audibility classifier using our self-collected dataset, which contains two categories: audible, and non-audible. This data set comprises 10k captions generated by ChatGPT, with an equal distribution of 5k captions labeled as audible and 5k labeled as non-audible, see Sec.~\ref{Audibility classifier}

\noindent\textbf{Implementation details\quad} \label{Implementation details}
To validate the effectiveness of our approach, we integrate our novel guided decoding technique with the zero-shot audio captioning model, NoAudioCaptioning \cite{deshmukh2024training}.

Our pipeline is implemented on a single Titan X GPU, utilizing a single beam search for caption generation. The system evaluates 512 candidate tokens within approximately 2 seconds per token, with a target sequence length set to 30 tokens.

The classifier $h_a$ in equation~\ref{eq:one} is based on the DistilBERT architecture \cite{sanh2019distilbert}. The optimization process employs the AdamW algorithm, configured with a batch size of 64 and a learning rate of 0.0003, over 40 epochs. A learning rate scheduler is applied to reduce the learning rate by a factor of 10 every 10 epochs.

We evaluate our pipeline using standard captioning metrics BLEU~\cite{papineni2002bleu}, METEOR~\cite{banerjee2005meteor}, ROUGE-L~\cite{lin2004rouge}, SPICE~\cite{anderson2016spicesemanticpropositionalimage}, CIDEr~\cite{cider}, semantic metrics CLAP-S~\cite{clap} for audio-caption alignment and BERT-S~\cite{bertscore} for semantic similarity with BERT embeddings, and introduce Audibility Accuracy (Aud) using our pre-trained audibility classifier (see Sec.~\ref{Audibility classifier}).

\noindent\textbf{Baselines\quad}
We compare our method with two fully supervised audio captioning models: Prefix AAC \cite{kim2023prefix} and RECAP \cite{ghosh2024recap}. Prefix AAC utilizes prefix tuning to refine the captioning process, while RECAP incorporates retrieval-augmented strategies to improve the quality of generated captions. Both models are evaluated in a out-of-domain audio captioning context and are trained without the use of additional data.

In addition, we compare our approach with several zero-shot audio captioning models: ZerAuCap \cite{salewski2023zeroshot}, NoAudioCaptioning \cite{deshmukh2024training}, WSAC \cite{kouzelis2023weakly}, and Zhang et al \cite{zhang2024zero}. ZerAuCap employs audio-language model guidance to generate captions in a zero-shot setting. On the other hand, NoAudioCaptioning, WSAC and Zhang et al generate captions without the need for audio data during training.

\noindent\textbf{Out-of-domain Audio Captioning\quad}\label{Out-of-domain Audio Captioning}
To ensure a fair zero-shot evaluation, we conduct our experiments within an out-of-domain framework, where training and test sets are drawn from different benchmark datasets. The model is trained on the Source benchmark without exposure to data from the Target benchmark, reflecting real-world applications where audio from the Target domain is often unfamiliar.

For example, when evaluating the NoAudioCaptioning baseline \cite{deshmukh2024training} on the Clotho test set \cite{drossos2020clotho}, the model is trained exclusively on the AudioCaps training set \cite{kim2019audiocaps}, and vice versa.

Notably, the out-of-domain setting does not apply to the ZerAuCap baseline \cite{salewski2023zeroshot}, as this method requires no training. In contrast, the other zero-shot methods use only textual data for training, while fully supervised methods rely on paired audio-text data.

\noindent\textbf{Results\quad}
The evaluation results from our pipeline on the AudioCaps and Clotho datasets are presented in Tab.~\ref{cross_domain_table}.

As shown, applying our method to the NoAudioCaptioning baseline results in improvements across all supervised metrics, achieving state-of-the-art performance in most cases when compared to other zero-shot methods, as well as to supervised methods in out-of-domain scenarios. These findings underscore the effectiveness and robustness of our approach.

\noindent\textbf{Ablation study\quad}
Standard deep learning approaches train image-conditioned language models by maximizing textual similarity with reference captions \cite{vinyals2015tellneuralimagecaption, xu2016showattendtellneural, Steven2017, anderson2018bottomuptopdownattentionimage}. However, public datasets often focus on the most salient objects, leading to models generating less distinctive captions that overlook finer details \cite{mscoco, flickr30k}.

To address this issue, we evaluated our method using two pre-trained image captioning models: (i) ZeroCap~\cite{tewel2022zerocap}, which operates in a zero-shot setting, and (ii) CLIPCap~\cite{mokady2021clipcap}, which functions in a fully supervised context.

We trained our classifier on a dataset that includes both finer-detailed captions and "regular" captions. The data collection and classifier training processes were carried out in a manner analogous to the approach used for the audibility classifier and dataset, as described in Sec.~\ref{Audibility classifier},~\ref{Implementation details}.

The results, as presented in Table~\ref{ablation_table}, indicate that our approach substantially improves the performance of the image captioning models, improving all supervised metrics, as well as the unsupervised metric, CLIP-S\cite{hessel2021clipscore}. This improvement in CLIP-S reflects the increased detailedness of the captions, as it measures the semantic alignment between the generated captions and the corresponding images. These findings highlight the potential of integrating pre-trained models with task-specific classifiers to create a versatile and powerful framework that can advance captioning systems across a wide range of tasks and modalities.

\section{Limitations} 
Despite the improvements of our framework, several limitations persist. First, relying on pre-trained models like GPT-2 may introduce biases that affect performance and generalization. Additionally, inference-time optimization adds computational overhead, making it less suitable for real-time or resource-constrained environments.

\section{Conclusion}
We propose a modality-agnostic framework that adapts captioning networks to the semantics of different distributions or contexts through classifier-driven guidance, focusing on audibility in zero-shot audio captioning. By combining pre-trained language models with an audibility classifier, our method significantly outperforms baseline models in zero-shot settings. 

Future work will explore applying this framework to additional modalities, such as video, and investigate methods to reduce optimization overhead.

\section{Acknowledgments}
This work was supported by a grant from the Tel Aviv University Center for AI and Data Science (TAD), and by the Israeli Ministry of Science, Israel-Singapore binational grant 207606.

\end{document}